\documentclass[a4paper,UKenglish,cleveref, autoref, thm-restate]{lipics-v2021}

\hideLIPIcs  

\usepackage{float}
\usepackage{subcaption}
\usepackage{booktabs}
\newcommand{\apex}[1]{``#1''}

\title{A Schema-aware Logic Reformulation for Graph Reachability} 

\author{Davide {Di Pierro}\footnote{Corresponding Author}}{Università degli Studi di Bari, Italy}
{davide@dipierro@uniba.it}{https://orcid.org/0000-0002-8081-3292}{}
\author{Stephan Mennicke}{Knowledge-based Systems Group, Technische Universität Dresden, Germany}{stephan.mennicke@tu-dresden.de}{https://orcid.org/0000-0002-3293-2940}{}

\author{Stefano Ferilli}{Università degli Studi di Bari, Italy}{stefano.ferilli@uniba.it}{https://orcid.org/0000-0003-1118-0601}{}

\authorrunning{D. {Di~Pierro}, S. Mennicke and S. Ferilli} 

\Copyright{D. {Di~Pierro}, S. Mennicke and S. Ferilli} 



\category{} 

\relatedversion{} 

\nolinenumbers 

\EventEditors{N. Musliu, H. Verhaeghe}
\EventNoEds{2}
\EventLongTitle{The 23rd workshop on Constraint Modelling and Reformulation}
\EventShortTitle{ModRef 2024}
\EventAcronym{ModRef}
\EventYear{2024}
\EventDate{September 02, 2024}
\EventLocation{Girona, Spain}
\EventLogo{}
\SeriesVolume{}
\ArticleNo{16}

\begin{document}

\maketitle

\begin{abstract}
Graph reachability is the task of understanding whether two distinct points in a graph are interconnected by arcs to which in general a semantic is attached. Reachability has plenty of applications, ranging from motion planning to routing. Improving reachability requires structural knowledge of relations so as to avoid the complexity of traditional depth-first and breadth-first strategies, implemented in logic languages. In some contexts, graphs are enriched with their schema definitions establishing domain and range for every arc. The introduction of a schema-aware formalization for guiding the search may result in a sensitive improvement by cutting out unuseful paths and prioritising those that, in principle, reach the target earlier. In this work, we propose a strategy to automatically exclude and sort certain graph paths by exploiting the higher-level conceptualization of instances. The aim is to obtain a new first-order logic reformulation of the graph reachability scenario, capable of improving the traditional algorithms in terms of time, space requirements, and number of backtracks. The experiments exhibit the expected advantages of the approach in reducing the number of backtracks during the search strategy, resulting in saving time and space as well.
\end{abstract}

\section{Background and Introduction}
Search in Artificial Intelligence has been one of the first hot topics in the field. For decades, researchers have endeavoured themselves in order to limit the computational effort for this task. Historically, the main search proposals were based on depth-first \cite{tarjan1972}, breadth-first \cite{bundy1984}, or A-star algorithms \cite{candra2020}. 

\begin{definition}
[Graph]
\label{def:graph}
A graph G is a pair (V, E) in which V represents the set of vertices and E is a relation of \textit{incidence}, associating vertices with each other. Each edge associates a vertex called the \textit{start} with a vertex called the \textit{end} of the edge. It can be indicated as $s \xrightarrow{e}o$, standing for an edge named e between the start s and the end o.
\end{definition}
Given a graph, we are able to define the \textit{reachability} in terms of the \textit{graph} definition:

\begin{definition}
[Graph Reachability]
\label{def:reachability}
Given a graph (V, E), a node $o$ is reachable from a node $s$ iff $\exists$ a path $p=e_0, e_1, ..., e_{k-1}$ s.t. $s \xrightarrow{e_0}s_1 \xrightarrow{e_1}s_2 \xrightarrow{...}s_{k-1} \xrightarrow{e_{k-1}}o$, and we call k the cardinality of the set $\{e_0, e_1, ..., e_{k-1}\}$ one of the lengths between $s$ and $o$. If a path $p$ does not exist, we say that $o$ is not reachable from $s$ and the distance between $s$ and $o$ is $\infty$.
\end{definition}
In general, we require $s_0, s_1, ..., s_{k-1}$ to be distinct to avoid useless loops, but this is negligible for the purpose of the work.

\noindent
These definitions are quite standard. 
In our setting, we will refer to \textit{Labelled Property Graphs (LPGs)} \cite{dipierro2023}, which are one of the most comprehensive designs for graph modelling and are supported by the most common graph database manager, Neo4j\footnote{https://neo4j.com/}. Throughout the paper, we will use the term \textit{graph} always referring to the LPG model.

\begin{definition}
[Labelled Property Graph]
\label{def:lpggraph}
Given an alphabet $\Sigma$, an LPG graph G is a 4-tuple (V, E, L, P) in which V and E are the same as for definition \ref{def:graph}, L is a labelling function $(V \cup E) \rightarrow 2^{\Sigma^{*}}$ associating to each node (resp. edge) a set of labels generated from the alphabet, and P is a named property function (V $\cup$ E) $\rightarrow$ ($\epsilon \cup \Sigma^{+}$) associating the value of a property to each node (resp. edge).  
\end{definition}

We have stressed the $\epsilon$ value to indicate that a property may be not present for a node (resp. edge). In current AI solutions, graphs are gaining momentum thanks to their flexibility, performance, and not to mention interpretability, an extremely desired feature nowadays. In opposition to all traditional databases (e.g. relational ones), graph databases do not have inherent support for schemas of data. Schemas of databases govern the structure of data including (internal and external) constraints, types, and properties. The schema-less nature of graph databases has a positive impact on data integration, merging, and mapping. On the other hand, the absence of a common vocabulary and conceptualization gives rise to defeats in the process of integrating different sources due to data ambiguity, redundancy, and synonyms. Especially in the field of the \textit{Semantic Web (SW)}, graphs have been enriched with ontological information, providing a sort of schema for them. However, this information cannot be considered a schema since it does not prevent inconsistencies or any kind of protection, which is instead one of the most attractive properties of databases. Abiding with databases, there is still no support for schema creation in graph databases. Neo4j provides little support for the creation of small constraints like uniqueness or mandatory properties \cite{pokorny2017}, but we claim there is room for more potentialities in bridging the gap between schemes and graph structures. Since schemes for graphs are not defined yet, no standard is available. Hence, we provide a definition here for what we refer to as graph schemes.

\begin{definition}
[Graph Schema]
\label{def:graphschema}
A schema graph is a 4-tuple (E, R, P, S). E is the set of \textit{entities}. [...] R is the set of \textit{relationships}, a set of labelled functions E $\rightarrow$ $2^E$, expressing dependencies among instances of the involved entities. [...] Given an alphabet $\Sigma$, P is a named property function (E $\cup$ R) $\rightarrow$ $\{Int, String, Date\}$ associating to entities (resp. relationships) properties with their associated domains. [...] S is the \textit{subclass} relation E $\rightarrow$ E indicating that the first entity is a specialization of the latter. [...] Given an e $\in$ E, we define $\Gamma^0(e) = e$, $\Gamma^1(e) = S(e)$, and in general $\Gamma^{n}(e) = S(\Gamma^{n-1}(e))$ until $Thing$ is reached. We define $Super(e) = \bigcup \Gamma^n(e)$ as the fix point operator of $\Gamma$ applied to e to indicate the labels of e and all its super-entities.
\end{definition}

\begin{definition}
[Graph Schema]
\label{def:graphschema}
A schema graph is a 4-tuple (E, R, P, S). E is the set of \textit{entities}. A special entity \textbf{Thing} is always available. R is the set of \textit{relationships}, a set of labelled functions E $\rightarrow$ $2^E$, expressing dependencies among instances of the involved entities. It can be indicated as $s \xrightarrow{e}o$, standing for a relationship named e between the start entity s and the end entity o.  Given an alphabet $\Sigma$, P is a named property function (E $\cup$ R) $\rightarrow$ $\{Int, String, Date\}$ associating to entities (resp. relationships) properties with their associated domains. Without losing generality, we can refer to only integer, string and date values. S is the \textit{subclass} relation E $\rightarrow$ E indicating that the first entity is a specialization of the latter. Multiple inheritance is not supported; hence, S creates a tree structure, having entity \textbf{Thing} as root. Given an e $\in$ E, we define $\Gamma^0(e) = e$, $\Gamma^1(e) = S(e)$, and in general $\Gamma^{n}(e) = S(\Gamma^{n-1}(e))$ until $Thing$ is reached. We define $Super(e) = \bigcup \Gamma^n(e)$ as the fix point operator of $\Gamma$ applied to e to indicate the labels of e and all its super-entities.
\end{definition}

\noindent
Slightly changing Definition \ref{def:reachability}, we can define reachability for graph schemas.
\begin{definition}
[Graph Schema Reachability]
\label{def:schemareachability}
Given a graph schema (E, R, P, S), an entity $o$ is reachable from an entity $s$ iff $\exists$ a path $p=e_0, e_1, ..., e_{k-1}$ s.t. $s = s_0 \xrightarrow{e_0}s_1 \xrightarrow{e_1}s_2 \xrightarrow{...}s_{k-1} \xrightarrow{e_{k-1}}s_k = o$ s.t. $\forall i \in \{0, k-1\}$ $Super(s_i) \xrightarrow{e_i}Super(s_{i+1}) \in R$ and $k$ is one of the lengths between $s$ and $o$.
\end{definition}
Differently from Definition \ref{def:reachability}, schemas takes into account hierarchies. Given a graph, we need to define when a graph is consistent with respect to a graph schema.

\begin{definition}
[Graph Schema Compliance]
\label{def:graphschemacompliance}
Given a (labelled property) graph (V, E, L, K) and a graph schema (C, R, P, S), a graph is compliant with a graph schema iff:
\begin{itemize}
    \item $\forall v \in V$ $L(u) \subset C$ (i)
    \item $\forall e \in E$ $e \subset R$ (ii)
    \item $\forall v \in V$ $\forall k \in K(v)$ s.t. $K(v) \neq \epsilon$  $\implies$ $\exists p \in P$ $\exists l \in L(v)$ s.t. $p(l) \neq \epsilon$ $\wedge$ k and p are equally named (iii)
    \item $\forall e \in E$ $\forall k \in K(e)$ s.t. $K(e) \neq \epsilon$ $\implies$ $\exists p \in P$ $\exists r \in R$ s.t. $p(l) \neq \epsilon$ $\wedge$ k and p are equally named (iv)
    \item $\forall s \xrightarrow{e}o \in E$ $\implies$ $\exists r \in R$ s.t. $(ls, lo) \in r$ $\wedge$ $ls \in Super(s)$ $\wedge$ $lo \in Super(o)$ $\wedge$ e and r are equally named (v).
\end{itemize}
\end{definition}

\noindent
In short, conditions (i) and (ii) state that labels of nodes and edges \textit{must} exist in the schema. Conditions (iii) and (iv) state that every property of a node (resp. edge) in the graph \textit{must} be presented in the schema, and vice versa every property in the schema \textit{can} be present in the graph. Condition (v) states that for each edge in the graph, there \textit{must} be a relationship in the schema defined for the labels of the connected nodes, or one of their respective super-entities. Conversely, if a relationship exists in the schema it \textit{may} be represented in the graph.

The basis of the following considerations starts from this result:
\begin{lemma}[Reachability]
\label{lemma:reachability}
Given a graph G compliant with a graph schema S = (E, R, P, S), a node $o$ is reachable from node $s$ only if the label of $o$ is reachable from the label of $s$ in the schema. 
\end{lemma}

\begin{proof}
Suppose $\exists$ a path $p = s \xrightarrow{e_0}s_1 \xrightarrow{e_1}s_2 \xrightarrow{...}s_{k-1} \xrightarrow{e_{k-1}}s_k$ then, from (v) of Definition \ref{def:graphschemacompliance}, 
$\exists$ $c_0 \xrightarrow{r_0}c_1 \xrightarrow{r_1}c_2 \xrightarrow{...}c_{k-1} \xrightarrow{r_{k-1}}c_k \in R$ 
s.t. $c_0 \in Super(s)$, $c_1 \in Super(s_1)$, ..., $c_k \in Super(s_k)$ $\implies$ $c_k$ is reachable from $c_0$.
\end{proof}

In this work, we propose the use of graphs in combination with graph schema to facilitate graph-based algorithms, specifically the \textit{reachability} task. The proposed setting is implemented following the GraphBRAIN framework.

\section{GraphBRAIN}
GraphBRAIN (GB)\cite{ferilli2020} is a framework for knowledge graph management and fruition. It combines an LPG graph with graph schemas, following the provided definition. Small variations are applied to GraphBRAIN. Specifically, nodes are provided with exactly one label, even though Neo4j does not have this limitation; GB schemas also provide the possibility of expressing \textit{mandatory} properties for nodes (resp. edges) belonging to a specific entity (resp. relationship). With respect to definition \ref{def:graphschema}, P is a named property function (E $\cup$ R) $\rightarrow$ ($\epsilon \cup \Sigma^{+}$) $\times$ ${\{false, true}\}$ in which the boolean value specifies whether it is mandatory or not. With respect to the definition \ref{def:graphschemacompliance} of compliance, a new condition is introduced:
\begin{itemize}
    \item $\forall p = (C, \_, true) \in P$, $L(e) = C \implies \exists k = K(e)$ s.t. $k \neq \epsilon$  $\wedge$ k and p are equally named.
\end{itemize}
This condition states that for every mandatory property for an entity (neglecting its type) in the schema, there \textit{must} exist the same property for nodes belonging to the entity.
Finally, instead of yielding a single complete schema for the graph, we split the information into several schemas, which can be combined. Every schema provides a separate \textit{view} of the graph, each of them concerning a separate domain of use. Some of the domains regard computing (related to the history of computing, combining hardware and software knowledge), tourism (knowledge related to infrastructures, means of transport, attractions, points of interest, etc.), food (traditional food information), linguistics (multilingual knowledge to be used for linguistic tasks, e.g. for diachronic analysis of terms), and education (knowledge for the intelligent fruition of learning objects).

\noindent
GraphBRAIN is available in the form of an API \cite{ferilli2022}, providing also support for data mapping into other standard logic formalisms (e.g. OWL/RDF, Prolog). One of the main characteristics of GraphBRAIN is the capability to move between different formalisms in order to be compliant with several standards and take advantage of the reasoning processes available when moving between other data representation standards. As an example, when mapping GB instances and schemas onto OWL/RDF triples, we can introduce reasoners publicly available in the Semantic Web field \cite{mishra2011}. In this work, we propose the mapping onto a first-order logic formalism for both instances and schema, used for different purposes. The GB API is provided with mechanisms to translate both the schema and instances into Prolog-like facts, that can be interpreted by Prolog and Answer Set Programming \cite{lifschitz2019} programs. The API supports at least two different types of instance translations: the first one is structured-based, in which the main predicates are represented as \apex{node} and \apex{arc}, while the second is more semantic and expresses the specific meaning of labels, properties and arcs. The two representations are formally interchangeable and we can obtain the latter from the first and vice versa, the choice is only guided by convenience and by the easiness of expressing relevant predicate rules for the purpose. While the latter looks prominent when logically connecting pieces of information, in this work we prefer the first representation since we are interested only in the structure of connections in the schema, that will affect the representation in the instances, as it will be presented in the next section.

\section{Problem Reformulation}
Here we propose how to solve the problem in a more efficient way by resorting the lists in the \textbf{arcs/2} predicates. We frame the problem under the condition of having a graph and a graph schema expressed as first-order logic clauses. In general, we deal with Horn clauses \cite{angluin1992}, and the information about the graph and its schema is expressed as \textit{facts}, a specific Horn clause with no negated terms. As previously said, we prefer a structural representation of the schema, focusing on entities, relationships and the subclass relation. Specifically, we have the \textbf{entity/1}, \textbf{subclassOf/2}, and \textbf{arc/2} predicates. We report in Listing \ref{lst:exampleschemafacts} an excerpt of predicates involving the entity \textbf{Person}. It is not a top entity since its super-entity is \textbf{Agent} which, on the other hand, is a top entity because is a child of \textbf{Entity}, the equivalent of \textit{Thing}. \textbf{Person} has one sub-class (\textbf{PersonUser}), and some arcs with entities like \textbf{Event}, \textbf{Organization} and \textbf{Place}. Remind that all the arcs of the super-entity \textbf{Agent} are also arcs for \textbf{Person}.

\begin{lstlisting}[caption={Extract of ``User''-related facts.},label=lst:exampleschemafacts,captionpos=t,float,abovecaptionskip=-\medskipamount,
    basicstyle=\small\ttfamily]
entity(agent).
subclassOf(agent, entity).
entity(person).
subclassOf(person, agent).
entity(personUser).
subclassOf(personUser, person).
arc(person, collection).
arc(person, event).
arc(person, organization).
arc(person, place).
...
\end{lstlisting}

\begin{lstlisting}[caption={Excerpts of facts.},label=lst:exampleschemafacts,captionpos=t,float,abovecaptionskip=-\medskipamount,
    basicstyle=\small\ttfamily]
node(1).
node(2).
node(3).
...
label(1, a).
label(2, b).
label(3, c).
...
arcs(1, [2]).
arcs(2, [2,1]).
arcs(3, [4]).
\end{lstlisting}

For each distinct pair of entities in the graph schema knowledge base, it must be computed the minimum distance between the pair. Our resolution, algorithmically described in Listing \ref{lst:distancecomputation}, has been implemented in \textit{Answer Set Programming} and is inspired by the distance-vector routing algorithm \cite{xu1997} that takes as input the set of pair of entities and returns a function specifying for each pair its distance. The computation of all distances between entities can be regarded as a preprocessing step since, unless ontological changes, these do not need to be recomputed when reachability involves different nodes. Ontological changes happen when the domains of use, users and/or management or law requirements change which is, fortunately, not frequent. After this preprocessing step, we are interested in filtering only pertaining distances, that are the ones involving the target node. The new graph navigation \textit{must} take into account distances between labels. Given lemma \ref{lemma:reachability}, we can exploit reachability among labels of nodes before navigating the graph. This may, in principle, (almost) always represent a benefit for the computation since labels are extremely fewer than nodes and the resorting (with pruning) of paths should increase the likelihood of finding the target node in a shorter time. The resorting of nodes prunes path for which the distance of labels is $\infty$, and the remaining (feasible) paths will be sorted according to the distance between labels. For instance, given the labels $S$, $A$, $B$ and $T$ (target), with $distance(A,T)=1$ and $distance(B,T)=2$, if a node labelled $S$ has connections with $A$ and $B$, it will prefer the path over $A$ since, in principle, it will reach a node equally labelled the target one shortly. How often this approximation is reliable will be a matter of discussion and it is, in fact, dependent on the dataset. 
\noindent
Instances are represented with two predicates: \textbf{node/2} and \textbf{arcs/2} where the first one represents the id of the node associated with its label, while the latter all the connections between ids. The second term of the predicate \textbf{arcs} is the list of adjacent nodes, and it is the list that needs to be sorted. We will call \textbf{improved\_arcs} the predicate derived by the sorting of connections with the label distances, and the solution is shown in Listing \ref{lst:improved_arcs}, as well as the improved reachability version.

\begin{lstlisting}[caption={Algorithm for distance computation},label=lst:distancecomputation,captionpos=t,float,abovecaptionskip=-\medskipamount,
    basicstyle=\small\ttfamily]
Input: Set<Pair> pairs
Output: Pair --> Int distance
compute_distances(pairs):
    for each pair in pairs:
        distance(pair) := find_minimum_distance(pair, [])
    return distance

Input: Pair pair, List<Int> visited
Output: Int min_distance
find_minimum_distance(pair, visited):
    min_distance := infty
    if pair[0] = pair[1]
        return 0
    else 
        for each y in arc(pair[0], y) and y not in visited:
            distance := 
            find_minimum_distance(y, pair[1], [y | visited])+1
            if distance < min_distance
                min_distance := distance
        return min_distance
            
\end{lstlisting}

\begin{lstlisting}[caption={Sorting arcs and improved reachability},label=lst:improved_arcs,captionpos=t,float,abovecaptionskip=-\medskipamount,
    basicstyle=\small\ttfamily]
distance_target(X, D) :- distance(T, X, D), target(T).

compare_order(<, X, Y) :- node(X, L1), node(Y, L2), OX < OY,
     distance_target(L1, OX), distance_target(L2, OY).
compare_order(>, X, Y) :- node(X, L1), node(Y, L2), OX >= OY,
    distance_target(L2, OY), distance_target(L1, OX).
    
has_distance(X) :- node(X, L), distance_target(L, _).

sort_arcs :- forall(arcs(Node, List),
        (
            include(has_distance, List, FilteredList),
            predsort(compare_order, FilteredList, SortedList),
            assertz(improved_arcs(Node, SortedList))
        )
    ). 
    
%improved reachability
improved_reach(X, X).
improved_reach(X, Y) :- !, improved_reach(X, Y, []).
improved_reach(X, Y, _) :- improved_arcs(X, L), member(Y, L).
improved_reach(X, Y, Visited) :- improved_arcs(X, L), 
    member(Z, L), Z \= Y, \+ member(Z, Visited), 
    improved_reach(Z, Y, [Z|Visited]), improved_arcs(X, L).
\end{lstlisting}

\section{Evaluation}
We tested our modified approach over two datasets: GraphBRAIN and the Twitter LPG dataset\footnote{https://github.com/neo4j-graph-examples/twitter-v2}. These two datasets have a very different complexity in the graph schema. The GB-selected ontology represents more than 140 entities and more than 90 arcs between them. On the other hand, the Twitter graph schema is composed of 6 entities and 13 arcs. Given the difference in their granularities, we can compare how much advantage is gained with respect to the level of detail of schemas. For GB, we tested several excerpts of the dataset. The rationale was to start from a set of representative nodes (ranked with PageRank), and then use a spreading activation function to grasp connected nodes from these central nodes. We extracted $\sim$5k nodes and $\sim$5k arcs per execution for GB, and $\sim$700 nodes and $\sim$700 arcs per execution for Twitter. We report here only results for a specific execution, since multiple runs of the same experiment did not lead to significant changes. Evaluation has been conducted by comparing the \textit{reachability} before and after optimization, and by measuring three percentages: number of executions where the improved version led to better performance, saved execution time and number of saved backtracks. The first percentage tells us how frequent is the advantage of using the improved version, the second gives us a glance at the value of the optimization, while the latter one is an objective measure, non-dependant on the specific performance of the algorithm and the machine used. Since each excerpt started from a central node, we computed \textit{reachability} from all the nodes to the central one. We tested GB and Twitter using SWI-Prolog\footnote{https://www.swi-prolog.org/}. All experiments have been conducted on an Intel(R) Core(TM) i7-1065G7 CPU @ 1.30GHz-1.50 GHz processor with 16GB of RAM.

\noindent
Table \ref{tab:results} shows the results for an instance of execution.

\begin{table}[tb] 
  \caption{Results}
  \label{tab:results}
  \centering
  \begin{tabular}{>{\centering\arraybackslash}p{3cm} >{\centering\arraybackslash}p{3cm} >{\centering\arraybackslash}p{3cm} >{\centering\arraybackslash}p{3cm}}
    \toprule
    \textbf{Dataset} & \textbf{\% Improved Time} & \textbf{\% Saved Time} & \textbf{\% Saved Backtracking} \\
    \midrule
    GraphBRAIN & \textbf{77} & \textbf{75.8} & 53.7 \\
    Twitter  & 68.5 & 36.5 & \textbf{59.2} \\
    \bottomrule
  \end{tabular}
\end{table}

The results demonstrated the usefulness of the approach, especially in the GB setting, where the graph schema is well-detailed and pruned to a greater extent. The second column tells us that, in 77\% of reachability computations, the improved version reduced the needed time to compute to fulfil the task. On average, the \% saved time, computed as 
$$\frac{time_{reachability} - time_{improved\_reachability}}{time_{reachability}} \cdot 100$$
is more than 75, also reaching peaks of 99\% for some cases. In terms of orders, the magnitude of saved time is in the order of 100 i.e., the improved version is 100x faster than the \textit{blind} version on average.
\noindent
Following our expectations, the same results do not happen for the Twitter dataset. However, it is still valuable to use a graph schema-aware reachability in more than 68\% of cases. Surprisingly, the percentage of saved backtracking is even higher than GB, even if the saved time is much lower. We have a guess for this behaviour: since there is high connectivity among nodes in Twitter and there are several arcs connecting nodes of the same class, our schema-aware reachability avoids useless loops visiting nodes of the same class and, at a certain point, still reaches the target node.

Times have been computed with the \textit{date} library of Prolog, considering the difference between the absolute time before and after the call to the \textit{reachability} query, as shown in Listing \ref{lst:dateprolog}.

\begin{lstlisting}[caption={Date library in Prolog},label=lst:dateprolog,captionpos=t,float,abovecaptionskip=-\medskipamount,
    basicstyle=\small\ttfamily]
measure_execution_time(Query, Time) :-
    get_time(Start),           % Get time before the query
    call(Query),               % Execute the query
    get_time(End),             % Get time after the query
    Time is End - Start.       % Calculate the elapsed time
\end{lstlisting}

\section{Related Work}
The reachability problem has roots in many computer science areas, especially in some graph-based structures that were pioneers of AI like Petri Nets \cite{czerwinski2020}. In general, this problem has been largely solved by database indexes \cite{zhang2023}; however, finding alternatives and more general solutions is still valuable given that not all graphs belong to a database and, perhaps, new solutions may be integrated with indexes as well. From the general reachability problem, a class of subproblems emerged with more specific purposes like the \textit{k-hop reachability} \cite{wang2023}, and the \textit{shortest path problem} \cite{marcucci2024}. 

Although historically there was a strict connection between logic and graph representation \cite{rensink2004,acclavio2020}, the problem of formalizing and solving the reachability problem with logic programs has been underrepresented. Heljanko et al. \cite{heljanko1999} formalized and solved the problem when designing deadlock detention, without introducing any specific optimization for reachability. A specific logic language \cite{cruz2014} has been provided for solving graph-based algorithms, but no optimized solutions were introduced in it. In the SW field, some restrictions can be overcome by introducing extensions with first-order logic for representing new graph rules \cite{magka2012}. With the introduction of uncertainty in logic programming, new applications become prone to be solved by logic inference. One of the most common regards recommendation with graphs \cite{catherine2016}. The absence of a common schema governing graph-based instances gave rise to the introduction of Inductive Logic Programming \cite{lavrac1994} techniques to infer it \cite{smole2011}. Modern graph database strategies may perform a satisfiability check in a much more efficient way, but in \cite{orejas2008} a logic for graph constraints definition was introduced. More interestingly, the types of constraints available in that work are quite similar to those currently available in Neo4j.

\section{Conclusions and Future Works}
In this work, we proposed a new approach to sorting graph connections in order to sensitively improve some tasks, \textit{reachability} in this specific case. Our experiments demonstrated the usefulness of this approach in first-order logic frameworks even in scenarios when the data schema does not seem to provide very peculiar information. We claim this approach is general, under the condition of having a specific heuristic function for which adjacent nodes can be sorted to improve a task. Further improvements of this work may take into account graphs that are not fully compliant with graph schemas, giving rise to solutions for handling knowledge incompleteness, extending the framework with statistical information (e.g., the likelihood of some paths), and introducing optimization over arcs that the algorithm may benefit from (e.g., cycles removal). 



\bibliography{lipics-v2021-sample-article}

\appendix

\end{document}